\documentclass{llncs}
\usepackage{amsmath}
\usepackage{dcolumn}
\usepackage{color}
\usepackage{hhline}
\usepackage{verbatim}
\usepackage{multirow}
\usepackage{amssymb}
\usepackage{rotating}
\usepackage{subfigure}
\usepackage{dblfloatfix}
\usepackage{enumitem}

\begin{document}
\title{BaitWatcher: A lightweight web interface for the detection of incongruent news headlines\thanks{This research article is published as a book chapter of \textit{Fake News, Disinformation, and Misinformation in Social Media-Emerging Research Challenges and Opportunities}. Springer, 2020.} 
}
\titlerunning{BaitWatcher}

\author{Kunwoo Park\inst{1} \and
Taegyun Kim\inst{2,3} \and
Seunghyun Yoon\inst{4} \and
Meeyoung Cha\inst{3,2} \and
Kyomin Jung\inst{4}}
\authorrunning{K. Park et al.}

\institute{Qatar Computing Research Institute, Doha, Qatar \and
Korea Advanced Institute of Science and Technology, Daejeon, Republic of Korea \and
Institute for Basic Science, Daejeon, Republic of Korea\and
Seoul National University, Seoul, Republic of Korea \\
\email{mcha@ibs.re.kr} \\
}

\maketitle
\begin{abstract}
In digital environments where substantial amounts of information are shared online, news headlines play essential roles in the selection and diffusion of news articles. Some news articles attract audience attention by showing exaggerated or misleading headlines. This study addresses the \textit{headline incongruity} problem, in which a news headline makes claims that are either unrelated or opposite to the contents of the corresponding article. We present \textit{BaitWatcher}, which is a lightweight web interface that guides readers in estimating the likelihood of incongruence in news articles before clicking on the headlines. BaitWatcher utilizes a hierarchical recurrent encoder that efficiently learns complex textual representations of a news headline and its associated body text. For training the model, we construct a million scale dataset of news articles, which we also release for broader research use. Based on the results of a focus group interview, we discuss the importance of developing an interpretable AI agent for the design of a better interface for mitigating the effects of online misinformation.
\end{abstract}

\section{Introduction}

The dissemination of misleading or false information in news media has become a critical social problem~\cite{kwon2013prominent}. Because information propagation online lacks verification processes, news contents that are rapidly disseminated online can put veiled threats to society. In digital environments that are under information overload, people are less likely to read or click on the entire contents; instead, they read only news headlines~\cite{gabielkov2016social}. A substantial amount of news sharing is headline-based, where people circulate news headlines without necessarily having checked the full news story. News headlines are known to play an essential role in making first impressions on readers~\cite{reis2015breaking}, and these first impressions have been shown to persist even after the full news content has been read~\cite{ecker2014effects}. Therefore, if a news headline does not correctly represent the full news story, it could mislead readers into the promotion of overrated or false information, which becomes hard to revoke.

\begin{figure}[t]
{
\centering
\includegraphics[width=0.7\columnwidth]{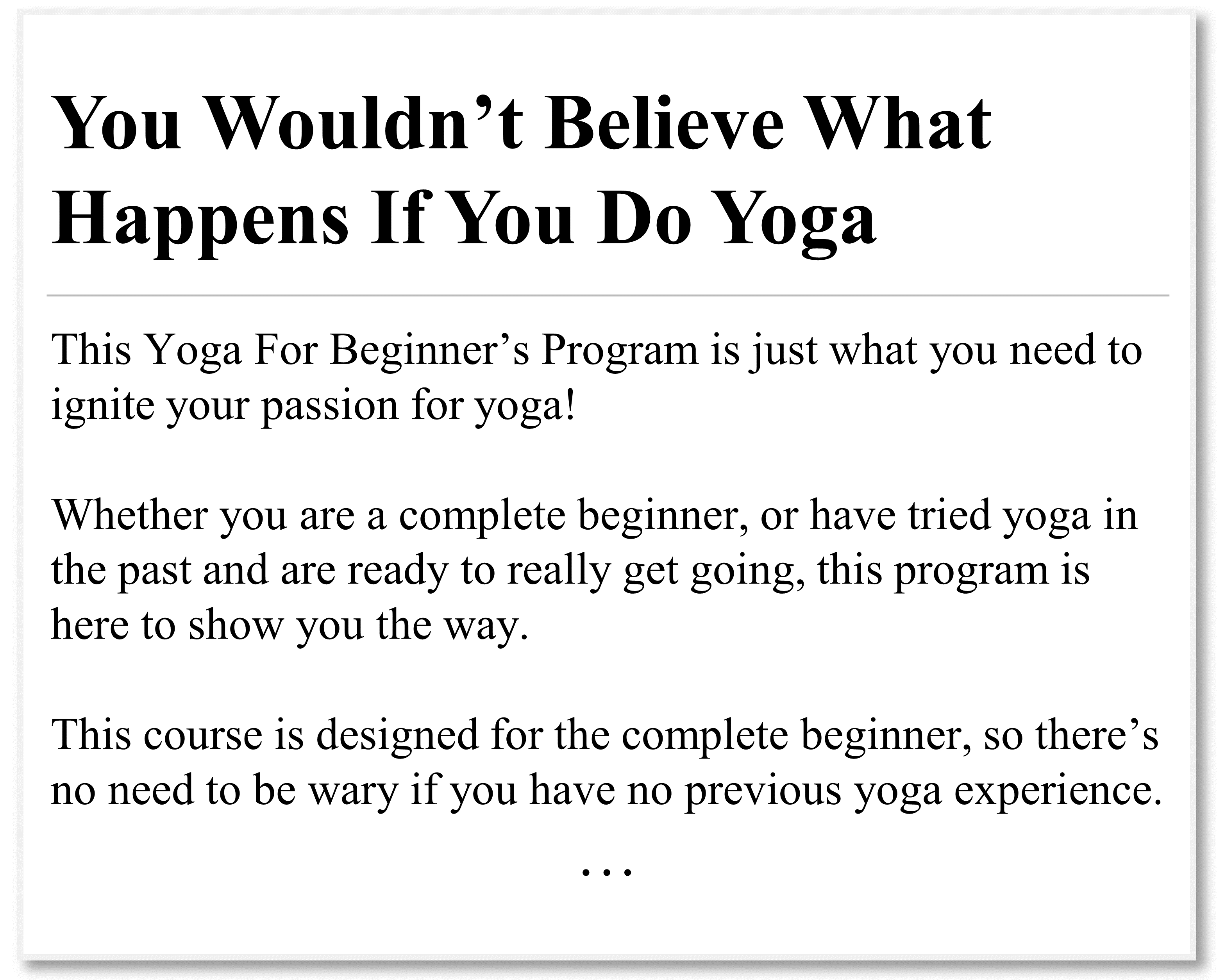}
\caption{
An example of news article with the incongruent headline.
}
\label{fig_example_design}
}
\end{figure}

This paper tackles the problem of headline incongruence~\cite{chesney2017incongruent}, where a news headline makes claims that are unrelated to or distinct from the story in the corresponding body text. Figure~\ref{fig_example_design} shows such an example, where the catchy news headline promises to describe the benefits of yoga, yet the body text is mainly an advertisement for a new yoga program. While this mismatch can be recognized when people read both the headline and the body text, it is almost impossible to detect it before clicking on the headline in online platforms. Incongruent news headlines make not only incorrect impressions on readers~\cite{ecker2014effects} but also become problematic when they are shared on social media, where most users share content without reading the full text~\cite{gabielkov2016social}. Therefore, the development of an automated approach that detects incongruent headlines in news articles is crucial.
Identifying incongruent headlines in advance will more effectively assist readers in selecting which news stories to consume and, thus, will decrease the likelihood of encountering unwanted information. 

Previous research has tried to detect misleading headlines by either analyzing linguistic features of news headlines~\cite{blom2015click,chen2015misleading} or analyzing textual similarities between news headlines and body texts~\cite{ferreira2016emergent,wang2017bilateral}. However, the lack of a large-scale public dataset hinders the development of sophisticated deep learning approaches that will be better suited for such challenging detection tasks, which typically require a million-scale dataset across various domains~\cite{lowe2015ubuntu,go2009twitter}. This study aims at filling this gap by proposing an automated approach for generating a million-scale dataset for headline incongruity, developing deep learning approaches that are motivated by hierarchical structures of news articles, and evaluating the model in the wild by developing a lightweight web interface that estimates the likelihood of an incongruent news headline.  


Our contributions are summarized as follows:

\begin{enumerate}
\item We develop a million-scale dataset for the incongruent headline problem, which covers almost all of the news articles that were published in a nation over two years. The corpus is composed of pairs of news headlines and body texts along with the annotated incongruity labels. The automatic approach for annotation can be applied to any news articles in any language and, therefore, will facilitate future research on the detection of headline incongruity in a broader research community.

\item We propose deep hierarchical models that encode full news articles from the word-level to the paragraph-level. Experimental results demonstrate that our models outperform baseline approaches that were proposed for similar problems. We extensively evaluate our models with real data. Manual verification successfully demonstrates the efficacy of our dataset in training for incongruent headline detection. 

\item To facilitate news reading in the wild, we present BaitWatcher --- a lightweight web interface that presents the prediction results that are obtained based on deep learning models \textit{before} readers click on news headlines. Along with this study, implementation details and codes will be shared. BaitWatcher is platform-agnostic; hence, it can be applied to any online news service. The results of a user study of focus group interviews not only support the effectiveness of the BaitWatcher web interface but also reveal a need for the development of interpretable models.

\end{enumerate}

\section{Related work}

\subsection{Learning-based approaches for detecting misleading headlines}

Interest has been growing in the automatic detection of misleading headlines. Previous studies constructed datasets that were annotated by journalists or crowd-sourced workers and proposed machine learning approaches. For example, a recent study suggested a co-training method for the detection of ambiguous and misleading headlines from pairs that consist of a title and a body text~\cite{wei2017learning}. We review the literature for each type of misinformation and its relation to the headline incongruence problem.

A series of studies have focused on the detection of \textit{clickbait headlines}, which are a type of web content that attracts an audience and induces them to click on a link to a web page~\cite{chen2015misleading}. There is no single and concise definition in the literature; clickbait is regarded as an umbrella term that describes many techniques for attracting attention and invoking curiosity to entice the reader to click on a headline~\cite{kuiken2017effective}. One study~\cite{chakraborty2016stop} released a manually labeled dataset and developed an SVM model for the prediction of clickbait based on linguistic patterns of news headlines. Using this dataset, other researchers suggested a neural network approach that predicts the clickbait likelihood~\cite{rony2017diving}. A national-level clickbait challenge was held, where the objective was to identify social media posts that entice readers to click on a link~\cite{clickbait_challenge}. The significant difference between clickbait and headline incongruence is that clickbait is characterized solely by the headline, whereas an incongruent headline defined by the relationship between the headline and the body text. These definitions are not mutually exclusive: a clickbait headline can also be incongruent with its main article. Clickbait headlines may be acceptable if they represent corresponding body texts accurately; however, the consequences can be more severe if catchy headlines mislead people with incorrect information.

The detection of headline incongruence is also related to the stance detection task, which aims at identifying the stance of specified claims against a reference text. The Emergent project~\cite{ferreira2016emergent} provides a dataset of 300 rumored claims and 2,595 associated news articles, each of which is labeled by journalists to indicate whether the stance of the article is \textit{for}, \textit{against}, or \textit{observing} the claim. The Fake News Challenge 2017 was held to promote the development of methods for estimating the stance of a news article~\cite{fnc1}. This dataset provides 50,000 pairs of headlines and body texts that were generated from 1,683 original news articles. Each data entry is annotated with one of the following four stances: \textit{agree}, \textit{disagree}, \textit{discuss}, and \textit{unrelated}. While many teams attempted to employ deep learning models (e.g., ~\cite{chopra2017towards,riedel2017simple}), the winning model was a simple ensemble approach that combines predictions from XGBoost~\cite{chen2016xgboost} that are based on hand-designed features and a deep convolution dual encoder that independently learns word representations from headlines and body texts using convolutional neural networks. 

Stance detection is technically similar to the headline incongruence problem in that they consider the textual relationship between a headline (claim) and the corresponding body text (reference text). It may be possible to transform three or four stances into binary classes, such as \textit{related} and \textit{unrelated}. But most available datasets cannot be directly utilized for the training of deep learning models for the headline incongruence problem because a \textit{related} headline can also be \textit{incongruent}. For example, the article in Figure~\ref{fig_example_design} would be labeled as related because both the headline and the body text cover the topic of yoga; however, the headline and the body text are incongruent.

This current study tackles the headline incongruence problem~\cite{chesney2017incongruent}, which is a significant kind of misinformation that originates from a discrepancy in a news headline and the corresponding body text. No million-scale dataset has been openly available for this problem.

\subsection{Designing web interfaces for news readers}

A line of research was conducted in which a news service was constructed as a separate system. A decade ago, a pioneering study provided a news service on the web~\cite{park2009newscube}. The researchers designed NewsCube, which was a news service that aimed at mitigating the effect of media bias. The service provided readers with multiple classified viewpoints on a news event of interest, which facilitated the formulation of more balanced views. Most recently, a study implemented a web system that highlights objective sentences in a user text to mitigate the biased reporting that facilitates polarization~\cite{lovering2018fact}. Another study developed a visualization tool that enables Twitter users to explore the politically-active parts of their social networks~\cite{gillani2018me} and conducted a randomized trial to evaluate the impact of recommending accounts of the opposite political ideology. The construction of a separate news service enables researchers to investigate the effects of a machine while controlling for other factors. However, it mostly serves as a proof of concept; hence, the impact on end-users is limited.

Many stand-alone systems suffer from gaining traffic. Therefore, other studies have developed a lightweight browser widget that operates with available news services on the web, which enables more users to be reached in practice. One study presented a browser widget that encourages the reading of diverse political viewpoints based on the selective exposure theory~\cite{munson2013encouraging}. According to a field deployment study, showing feedback led to more balanced exposure. A browser extension was also implemented to help people determine whether a headline is clickbait or a general headline~\cite{chakraborty2016stop}; however, the effects of this mechanism were not evaluated.

Motivated by these studies that employ browser widgets, this study implements a lightweight web interface that helps readers determine whether a specified headline is incongruent before clicking on the headline. We also conduct a user study via questionnaires and in-depth interviews to estimate the impact of the web interface.

\section{Data and Methodology}

This section presents the approach to building a million-scale dataset for the headline incongruence problem and the methodology for detecting such misleading headlines via neural networks. The objective is to determine whether a news article contains an incongruent headline, given a pair that consists of a headline and a body text. For detection models, we call the output probability of being an incongruent headline the \textbf{incongruence score}.

\subsection{Dataset generation}

One natural method for constructing a labeled dataset is for researchers and crowdsourced workers to manually annotate data. However, the training of sophisticated classification algorithms requires a large dataset, which is not feasible to obtain via manual annotation due to high cost and reliability issues. Alternatively, this work presents a systematic approach for the automatic generation of million-scale datasets that are composed of incongruent and correct headlines.

\begin{figure}[t]
{
\centering
\includegraphics[width=0.9\columnwidth]{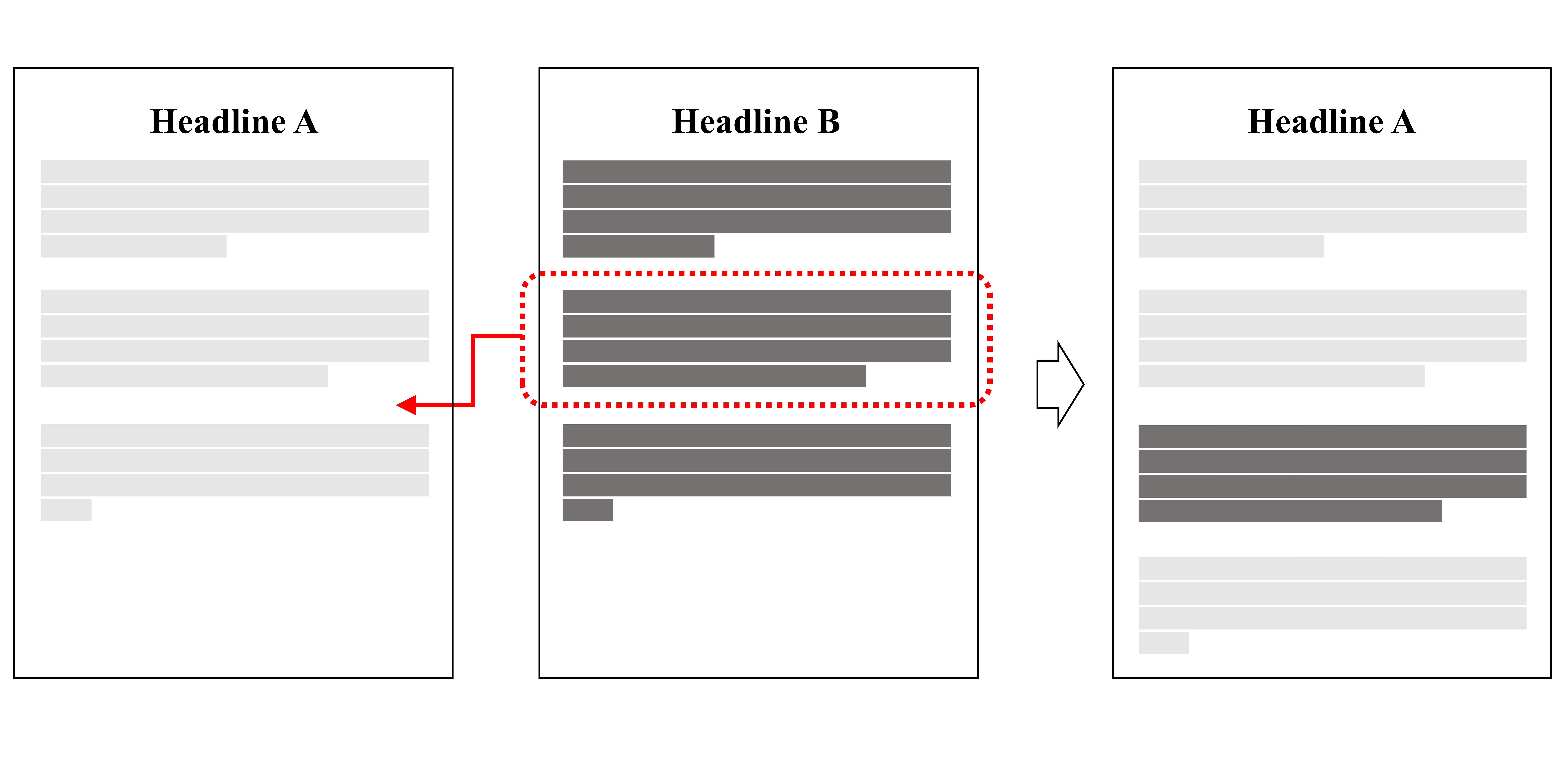}
\caption{
An illustration on the generating process of incongruent headlines. 
}
\label{fig_article_gen}
}
\end{figure}

First, we crawled a nearly complete set of news articles that were published in South Korea from January 2016 to October of 2017. From over 4 million news articles, we conducted a series of cleansing steps, such as removal of noncritical information (e.g., reporter name and nontextual information such as photos and videos). Next, we transformed word tokens to integers, which is released with vocabulary to help researchers utilize the dataset without being hindered by a language barrier.

To label the incongruity of headlines for millions of news articles, we implanted unrelated or topically inconsistent content into the body text of original news articles rather than crafting new headlines. Figure~\ref{fig_article_gen} illustrates the generation process of incongruent headlines. This process can produce a pair that consists of a headline and a body text such that the headline tells stories that differ from the body text content. 
Hence, the automation process for creating incongruent-labeled data involves the following steps: (1) sampling a target article from the corpora, (2) sampling part-of-content from another article of the corpora, and (3) inserting this part-of-content into the target article. We controlled the topics of the sampled paragraphs to be similar to each target article by employing the meta-information on news articles (e.g., news category).

We created the congruent-labeled data by selecting them from suitable corpora. No headline in this set overlaps with the incongruent-labeled data. Nonetheless, this process may yield false-negative instances if a real article that has an incongruent headline is chosen inappropriately as a target. We conducted additional steps to reduce the number of false negatives via rule-based processing, such as the inspection of advertising phrases with an n-gram dictionary. We also hired human annotators to read 1,000 randomly sampled articles from the created dataset and to check whether those articles are labeled correctly. These efforts minimize the number of errors that can arise from the automatic generation process. The final corpus is composed of a training set of 1.7 M news articles that are balanced against the incongruity label. For evaluation, we maintained separate development and test sets of 0.1 M instances each. The statistics of the datasets are presented in Table~\ref{table_dataset}.

\begin{table}[t]
\centering
\begin{tabular}{lrr}
 	 & \multicolumn{1}{c}{Mean} &
 	\multicolumn{1}{c}{Std. Error} \\\hline
Number of tokens in headline & 13.71 & 0.003 \\
Number of tokens in body text & 513.97& 0.208 \\
Number of paragraphs in body text & 8.17 & 0.004 \\
Number of tokens in paragraph & 61.7 & 0.018 \\\hline
\end{tabular}
\vspace{1mm}
\caption{Dataset statistics.}
\label{table_dataset}
\end{table}

This approach is language-agnostic; hence, it can be applied to any news corpora of any language. The generated dataset is publicly available on the GitHub page.\footnote{http://github.com/david-yoon/detecting-incongruity/} 


\subsection{Baseline approaches}

We introduce four baseline approaches that have been applied to the headline incongruence problem. Feature-based ensemble algorithms have been widely utilized for their simplicity and effectiveness. Among various methods, the XGBoost algorithm has shown superior performance across various prediction tasks~\cite{chen2016xgboost}. For example, in a recent challenge on determining the stances of news articles~\cite{fnc1}, the winning team applied this algorithm based on multiple features to measure similarities between the headline and body text~\cite{fnc_winner}. As a baseline, we implemented the \textbf{XGBoost (XGB)} classifier by utilizing the set of features that are described in the winning model, such as cosine similarities between the headline and body text. In addition to this model, we trained \textbf{support vector machine (SVM)} classifiers based on the same set of features.

\noindent\textbf{Recurrent Dual Encoder (RDE): } A recurrent dual encoder that consists of dual Recurrent Neural Networks (RNN) has been utilized to calculate a similarity between two text inputs~\cite{lowe2015ubuntu}. 
We apply this model to the headline incongruence problem via dual RNNs based on gated recurrent unit (GRU) that encode the headline and body, respectively.
When a RNN encodes word sequences, each word is passed through a word-embedding layer that converts a word index to a corresponding 300-dimensional vector.
After the encoding step, the probability of being incongruent headline is calculated by using the final hidden states of RNNs for headline and body text. The incongruence score in the training objective is as follows:

\begin{equation}
\begin{aligned}
& p(\text{label}) = \sigma ( ({h_{t_h}^H})^{\intercal}M~h_{t_b}^B + b ), \\
& \mathcal{L} = -\log \prod_{n=1}^{N} p(\text{label}_n | h_{n,t_h}^H, h_{n,t_b}^B),
\end{aligned}
\label{eq_de_loss}
\end{equation}
where $h_{t_h}^H$ and $h_{t_b}^B$ are last hidden state of each headline and body text RNN with the dimensionality $h \in \mathbb{R}^d$. The $M \in \mathbb{R}^{d \times d}$ and bias $b$ are learned model parameters. $N$ is the total number of samples used in training and $\sigma$ is the sigmoid function.

\noindent\textbf{Convolution Dual Encoder (CDE):} Following the Convolutional Neural Network (CNN) architecture for text understanding \cite{kim2014convolutional}, we apply Convolutional Dual Encoder to the headline incongruence problem. Taking the word sequence of headline and body text as input to the convolutional layer, we obtained a vector representation $\boldsymbol{v}= \{ v_{i} | i=1, \cdots, k\}$ for each part of the article through the max-over-time pooling after computing convolution with $k$ filters as follows:
\begin{equation}
\begin{aligned}
& v_i = g(f_i(W)),
\end{aligned}
\end{equation}
where $g$ is max-over-time pooling function, $f_i$ is the CNN function with \textit{i}-th convolutional filter, and $W \in \mathbb{R}^{t \times d}$ is a matrix of the word sequence.
We use dual CNNs to encode a pair of headline and body text into vector representations. After encoding each part of a news article, the probability that a given article has the incongruent headline is calculated in a similar way to the equation (\ref{eq_de_loss}).

\subsection{Proposed methods}

While the available approaches perform reasonably for short text data, dealing with a long sequence of words in news articles will result in degraded performance~\cite{pascanu2013difficulty}. 
For example, RNN that is utilized in RDE performs poorly in remembering information from the distant past.
While CDE learns local dependencies between words, the typical short length of its convolutional filter prevents the model from capturing any relationships between words in distinct positions. The inability to handle long sequences is a critical drawback of the standard deep approaches to the headline incongruence problem because a news article can be very long. As presented in Table~\ref{table_dataset}, the average word count per article is 513.97 in our dataset.

Therefore, we fill this gap by proposing neural architectures that efficiently learn hierarchical structures of long text sequences. We also present a data augmentation method that efficiently reduces the length of the target content while increasing the size of the training set.

\noindent\textbf{Hierarchical Recurrent Dual Encoder (HRDE): }
Inspired by previous approaches that models text using a hierachical architecture~\cite{zhou2016attention,yang2016hierarchical,yoon2018learning}, this model divides the text into a list of paragraphs and encodes the entire text input from the word level to the paragraph level using a two-level hierarchy of RNN architectures.

For each paragraph, the word-level RNN encodes the word sequences $\boldsymbol{w}_p={\{w_{p,1:t}\}}$ to $\boldsymbol{h}_p={\{h_{p,1:t}\}}$. 
Next, the hidden states of the word-level RNN are fed into the next-level RNN that models a sequence of paragraphs while preserving the order. The hierarchical architecture can learn textual patterns of news articles with fewer sequential steps for RNNs compared to the steps required for RDE. While RDE involves an average of 513.97 steps to learn news articles in our dataset, AHDE only accounts for 61.7 and 8.17 steps on average for word- and paragraph-level of RNN, respectively (see Table~\ref{table_dataset}). 
The hidden states of hierarchical RNNs are as follows:
\begin{equation}
\begin{aligned}
 &h_{p,t} = f_{\theta}(h_{p,t-1}, w_{p,t}), \\
 &u_p = g_{\theta}(u_{p-1}, h_p),
\end{aligned}
\label{eq_hrde}
\end{equation}
where $u_p$ is the hidden state of the paragraph-level RNN at the \textit{p}-th paragraph sequence, and $h_p$ is the word-level RNN's last hidden state of each paragraph $h_p \in\{h_{1:p,t}\}$.
We use the same training objective as the RDE model such that the incongruence score, the probability of having incongruent headlines, is calculated as follows:
\begin{equation}
\begin{aligned}
& p(\mbox{label}) = \sigma ( ({u_{p_h}^H})^\intercal M~u_{p_b}^B + b )
\end{aligned}
\label{eq_hrde_loss}
\end{equation}

\begin{figure}[t]
\small
\centering
\includegraphics[width=0.9\columnwidth]{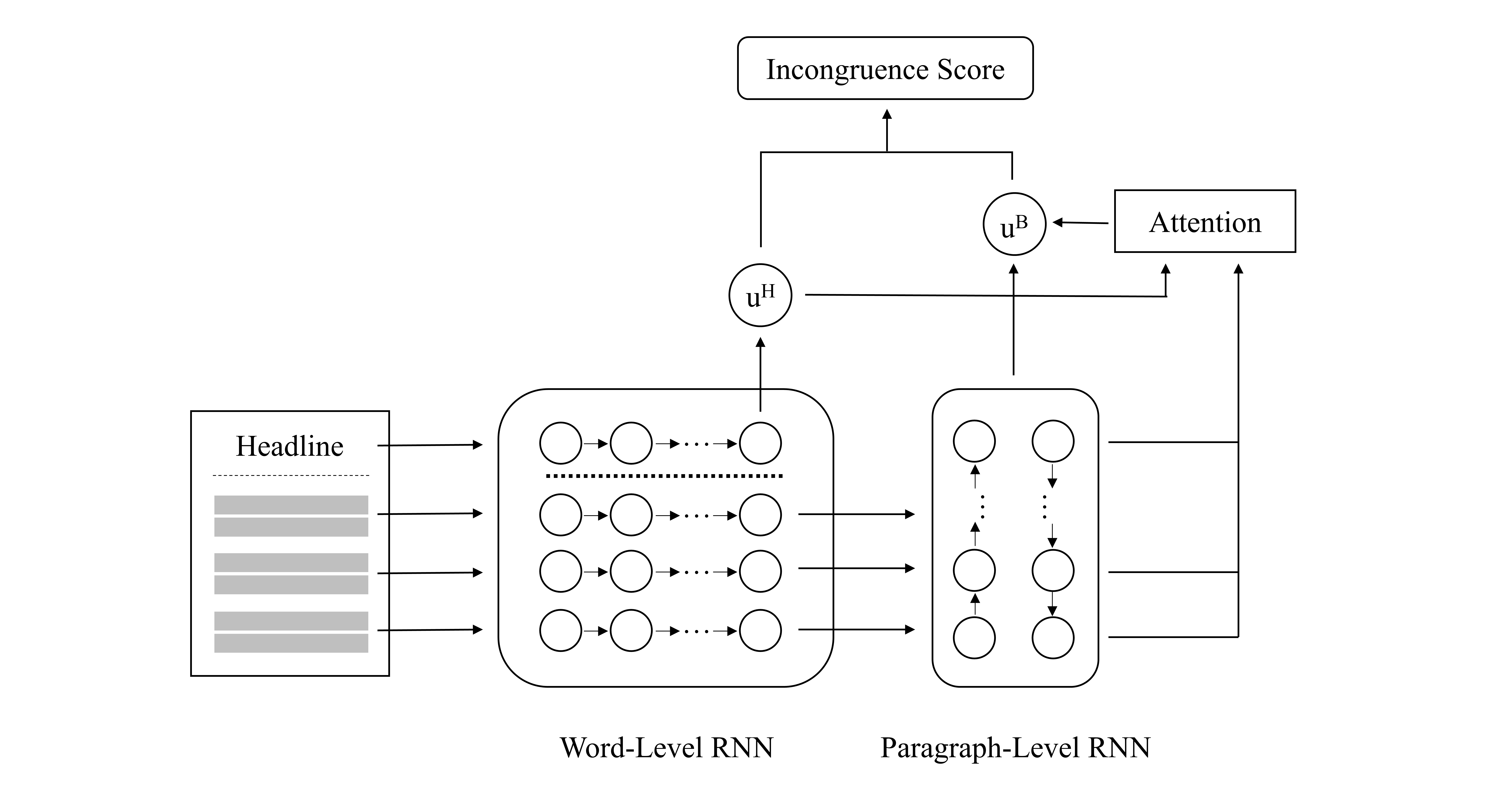}
\caption{
Diagram of AHDE. Every paragraph is encoded into a hidden state, and the sequence of the hidden states corresponding to each paragraph is further encoded into the hidden state corresponding to the entire body text. The model can learn importance of paragraphs in a body text for detecting headline incongruity from an attention mechanism.}
\label{fig_ahde}
\end{figure}

\noindent\textbf{Attentive Hierarchical Dual Encoder (AHDE): }
In addition to the hierarchical architecture of HRDE, attention mechanism is employed to the paragraph-level RNN to enable the model to learn the importance of each paragraph in a body text for detecting incongruity embedded in the corresponding headline. Additionally, we utilize bi-directional RNNs for the paragraph-level RNN to learn sequential information in both directions from the first paragraph and the last paragraph.

Figure~\ref{fig_ahde} illustrates a diagram of AHDE. Each $u_p$ of a body text is aggregated according to its correspondence with the headline as follows:
\begin{equation}
\begin{aligned}
 &s_p = \text{v}^\intercal tanh(\text{$W_u^B$}u_{p}^B + \text{$W_u^H$}u^H), \\
 &a_i = \text{exp}(s_i) / {\scriptstyle\sum_p}\text{exp}(s_p), \\
 &{u^B} = {\scriptstyle\sum_i} a_i u_{i}^B,
\end{aligned}
\label{eq_ahde}
\end{equation}
where $u_{p}^B$ indicates the \textit{p}-th hidden state of the paragraph-level RNN that learns the representation of a body text. The $u^H$ indicates the last hidden state of the paragraph-level RNN with the corresponding headline. Similar to HRDE, the incongruence score is calculated as follows:
\begin{equation}
\begin{aligned}
& p(\mbox{label}) = \sigma ( ({u^H})^\intercal M~u^B + b )
\end{aligned}
\label{eq_ahde_loss}
\end{equation}

\vspace*{1mm}
\noindent\textbf{Hierarchical Recurrent Encoder (HRE):}
The HRDE and AHDE model uses two hierarchical RNNs for encoding text from the word level to the paragraph level. Compared to non-hierarchical alternatives such as RDE and CDE, those models require higher computation resources in training and inference. Therefore, we investigate a moderate approach that models hierarchical structures of news articles with a simpler neural architecture. A body text is divided into paragraphs, each of which is represented by averaging word-embedding vectors of words within the paragraph. In other words, HRE calculates $h_{p}$ in equation (\ref{eq_hrde}) by the average of the word vectors in the paragraph $p$, $h_p = \sum_i embedding(w_i),$ $w_i\subset{\mbox{\textit{p}-th~paragraph}}$. Then, a paragraph-level RNN is applied to the paragraph-encoded sequence input, $h_{p}$, for retrieving the final encoding vector of the entire body text. The incongruence score is calculated by 

\begin{equation}
\begin{aligned}
& p(\mbox{label}) = \sigma ( ({h^H})^\intercal M~h_{p} + b )
\end{aligned}
\label{eq_hre_output}
\end{equation}
where $h^H$ indicates the average embedding vector of the words in the headline.

\subsubsection{\textbf{Independent Paragraph (IP) method}}

\begin{figure}[t]
\small
\centering
\includegraphics[width=0.9\columnwidth]{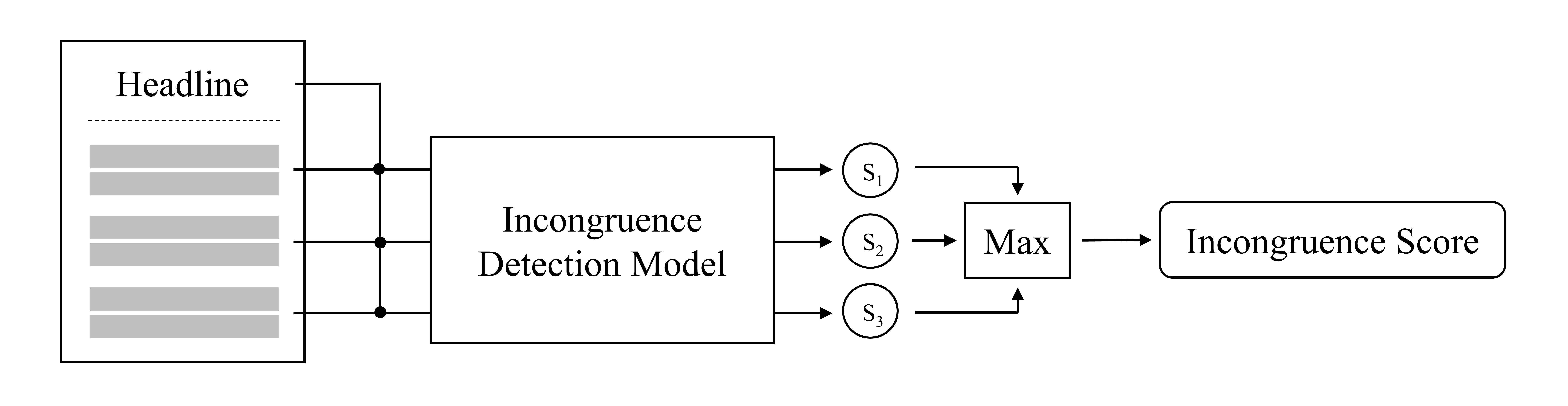}
\caption{
Diagram of the IP method. A body text is divided into multiple paragraphs, each of which is compared to the corresponding headline to calculate the incongruence score of each paragraph. The maximum value of all scores is the incongruence score for the pair of the headline and the body text.}
\label{fig_ip}
\end{figure}

In addition to the neural architecture, we propose a data augmentation method that divides a body text into multiple paragraphs and learns the relationship between each paragraph and the corresponding headline independently. For that purpose, we transformed every pair that consists of a headline and a body text in the original dataset into multiple pairs of the headline and each paragraph. This conversion process not only reduces the length of text that a model should process but also increase the total number of training instances. For example, the average number of words in a body text shrinks from 513.97 to 61.7 (see Table~\ref{table_dataset}), and the number of training instances increases from 1.7 M to 14.2 M. We expect that this difference makes the proposed deep learning models efficiently learn the pattern of the semantic mismatch between a headline and its body text. 

Figure~\ref{fig_ip} illustrates the diagram of the IP method, which computes \textbf{incongruence score} of each paragraph from its relationship with the corresponding headline. The final incongruence score for the pair of the headline and the body text is determined as the maximum of the incongruence score for the headline and each paragraph as follows:
\begin{equation}
\begin{aligned}
	& p(\mbox{label}) = max(s_{1:p}), \\
\end{aligned}
\label{eq_ip}
\end{equation}
where $s_p$ is the incongruence score calculated from the \textit{p}-th paragraph of the body text and the headline. 
The selection of the maximum score can better identify news articles that contain a paragraph that is highly unrelated to the news headline. We also tested other aggregation methods such as average and minimum, but max function led to the best performance.

With the use of the IP method,  hierarchical approaches consider sentence in a paragraph the lower unit in two-level hierarchy of neural architectures. In particular, the incongruence score of each detection model is calculated in the following ways:
\begin{itemize}
\item \textbf{XGB/SVM with IP:} For each paragraph, XGB/SVM measures the incongruence score by extracting features from its headline and the paragraph. 

\item \textbf{RDE/CDE with IP:} Both models encode word sequences in each paragraph of a body text and compare them with the corresponding headline.

\item \textbf{HRDE/AHDE with IP:} To obtain the incongruence score for each paragraph, the first-level RNN encodes word sequences for each sentence in the paragraph, and the second-level RNN takes as input the hidden states of the sentences that are retrieved from the first-level RNN.

\item \textbf{HRE with IP:} HRE calculates the mean of word vectors for each sentence. Then, a RNN encodes a sequence of sentences by taking the averaged word vectors as input.
\end{itemize}

\begin{table}[t]
\centering
\begin{tabular}{ccccc}
\hline
\multirow{2}{*}{Model} & \multicolumn{2}{c}{Without IP} & \multicolumn{2}{c}{With IP} \\ \cline{2-3} \cline{4-5} 
 	& Accuracy   &AUROC &Accuracy &AUROC\\ \hline 
SVM		&0.640	&0.703	&0.677	&0.809	\\
XGB		&0.677	&0.766	&0.729	&0.846	\\
CDE		&0.812	&0.9	&0.870	& 0.959	\\
RDE		&0.845	&0.939	&0.863	&0.955	\\
\hline
HRDE   &\textbf{0.885}  & \textbf{0.944}	&\textbf{0.881}	& \textbf{0.962}	\\
AHDE	&\textbf{0.904}	&\textbf{0.959}	&\textbf{0.895}	&\textbf{0.977}	\\
HRE		&0.85	&0.927	&0.873	&0.952	\\   \hline
\end{tabular}
\vspace{1mm}
\caption{
Model performance with and without the Independent Paragraph (IP) method. Top-2 scores are marked as bold. The top 4 rows indicate the baseline performance and the bottom 3 rows shows the performance of the proposed models.  
}
\label{table_accuracy_auroc}
\end{table}

\section{Evaluation Experiments}

\subsection{Automatic evaluation}

Table~\ref{table_accuracy_auroc} presents the performances of all approaches on the test set. We report the accuracy and the AUROC (area under the receiver operating characteristic curve) value, which is a balanced metric for the label distribution. Here, we make three main observations. 

First, among the baseline models, RDE realized the best performance with an accuracy of 0.845 and an AUROC of 0.939. The decent performance of RDE suggests that recurrent neural networks are well suited to the learning of sequential text representations of news articles, in contrast to the feature-based approaches and the convolutional encoders, which learn the local dependencies of word tokens. 

Second, the performance margin increased significantly when hierarchical structures were applied to RDE. In HRDE, the accuracy and AUROC increased by 0.04 and 0.05, respectively. Considering the hierarchical structure of news articles in the design of neural architectures may facilitate the learning of textual information of news articles more efficiently, such that headline incongruity can be more accurately identified. In contrast, in HRE, merely inputting the mean word representation for each paragraph into a single layer recurrent network did not yield a significant improvement. Compared to RDE, the accuracy increased with a margin of 0.005; however, the AUROC decreased by 0.012. Third, we found the attention ability of AHDE further enhanced the performance up to an accuracy of 0.904 and an AUROC of 0.959, namely, knowledge of relevant paragraphs in the detection of incongruent headlines facilitated the efficient examination of the relationship between the headline and each paragraph by AHDE. 

Last, the prediction performance increased significantly when the IP method was applied. RDE and CDE benefitted most from the application of the IP method; they even showed performances that were comparable to those of the hierarchical models. Although those simple models do not have a suitable structure for handling lengthy news data (on average, the body texts and the paragraphs contain 518.97 and 61.7 words, respectively, according to Table~\ref{table_dataset}), the IP method helped them examine the relationship between the headline and each paragraph more efficiently. 

\subsection{Manual evaluation}\label{manual_eval}

To test the efficacies of the dataset and the proposed models for the detection of incongruent headlines in the wild, we evaluated the pretrained models on more recently published news articles. We gathered 232,261 news articles that were published from January to April of 2018. Via evaluation of the model on this recently assembled dataset, we can measure the generalizability of our approaches to dataset generation and headline incongruity detection in practice.

First, we manually inspected random samples of news articles to determine whether they have incongruent headlines; however, we could not retrieve sufficiently many instances with incongruent headlines for evaluation. The lack of misleading articles is possibly due to the sparsity of such headlines in practice, despite their critical importance. Therefore, instead of manually labeling randomly sampled news articles, a majority of which may correspond to general headlines, we manually evaluated the top-$N$ articles in terms of the incongruence scores that are assigned by our model. Since models assign incongruence scores (output probabilities) based on their confidence for classification, we believe such evaluation successfully estimates the degree of precision of a prediction model. This type of assessment is widely used in tasks in which it is impossible to count all possible real cases in a dataset such as question answering system~\cite{ferrucci2012introduction}. 

\begin{figure}[t]
\small
\centering
\includegraphics[width=0.7\columnwidth]{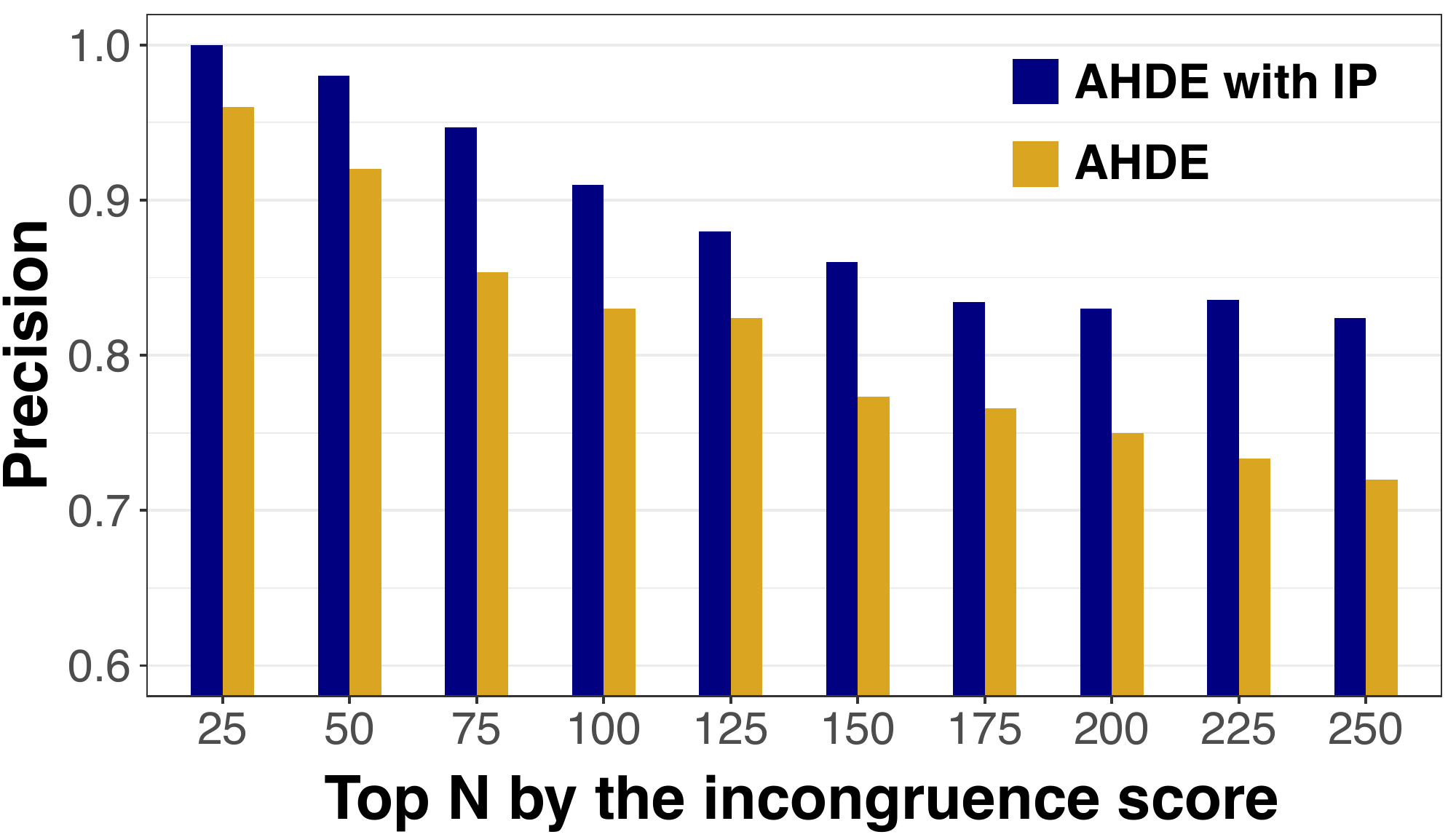}
\caption{
Precision values for detecting news articles with incongruent headlines in the newly gathered dataset. The x-axis shows the top-N articles by incongruence scores, and the y-axis presents its corresponding precision.
}
\label{fig_real}
\end{figure}

Figure~\ref{fig_real} presents the precision scores for the AHDE models that are trained with and without the IP method. The x-axis corresponds to the top-N articles in terms of the incongruence scores that are assigned by the models out of the newly gathered news articles over four months. The y-axis corresponds to the precision values of the top-N articles. Here, we make three observations. 

First, the AHDE model with the IP augmentation consistently shows higher precision than the AHDE model without the IP method. This finding supports the superior performance of the IP method across evaluations. Second, the AHDE model with the IP method realized a precision of 1.0 for the top 25 articles. Even though the model was trained on a separate dataset, it successfully filtered out real cases in which the headline conveys different information than the associated body text.
Third, when we evaluate the top 250 articles, the precision of the AHDE model with the IP method reduced to 0.82. This precision value is sufficiently high for the detection of incongruent headlines in real news platforms.

\section{BaitWatcher: A lightweight web interface for the detection of headline incongruity}

This section introduces a new web interface that aims at reducing the adverse effects of incongruent news headlines on the news reading experience. Incongruent headlines can mislead readers with an unexpected body text because they are one of the critical cues that are used in the selection of news articles in online environments. Before clicking on a headline and reading the body text, newsreaders are not able to determine in advance the content of the news story. We hypothesize that news readers will be empowered if they are given a choice and additional information about the headline incongruence score. As a proof of concept, we designed and implemented a web interface, namely, \textit{BaitWatcher}, that quickly reports the incongruence score. We conducted a focus group interview to investigate the effects of the web interface.

\subsection{Design and implementation}

\begin{figure}
\includegraphics[width=0.99\columnwidth]{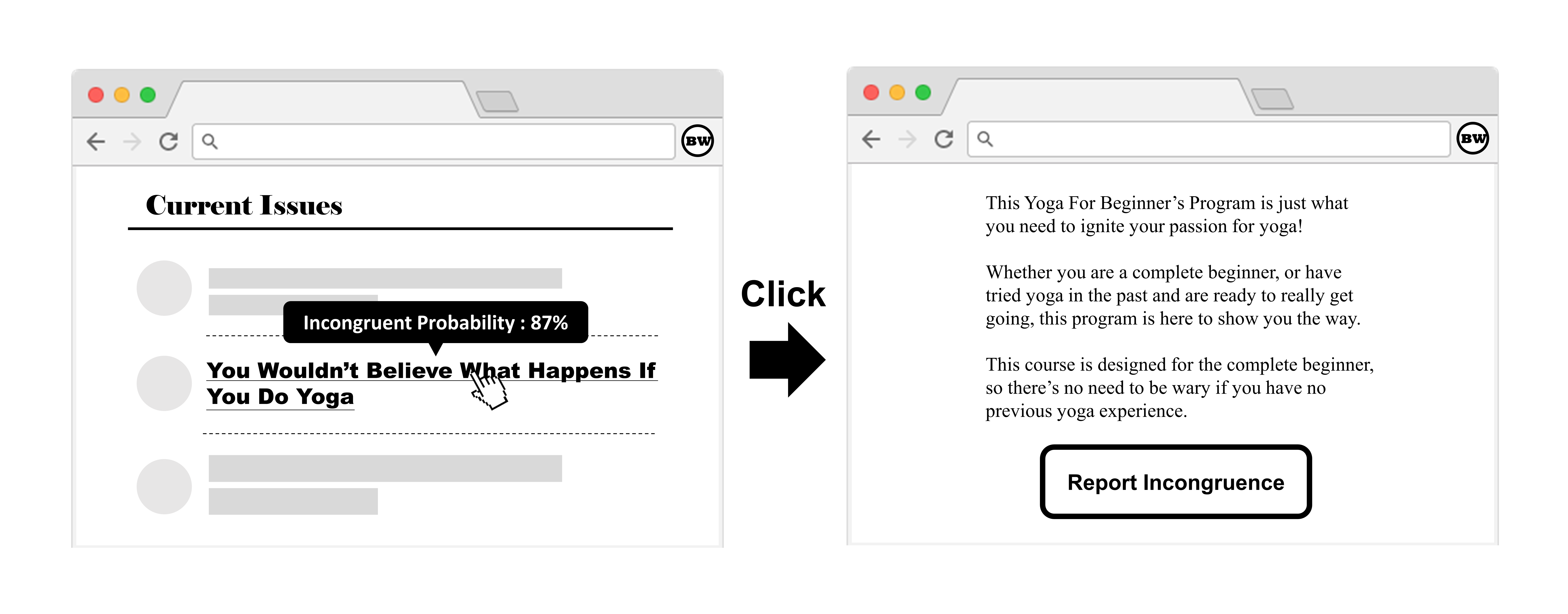}
\caption{The user interface of BaitWatcher.}
\label{bait_main}
\end{figure}

The main feature of BaitWatcher is that it reveals the likelihood of a specified news headline being incongruent to its full body text based on the presented deep learning model. This information is made visible to users \textit{before} they click on the headline to read the entire story. BaitWatcher is platform-agnostic and can be implemented on top of any news platforms. We expect that revealing the hidden information through a simple interface will empower news readers by helping them determine by themselves whether to read news articles with potentially incorrect headlines or not. As shown in Figure~\ref{bait_main}, if a user hovers a mouse pointer over a news headline of interest, the BaitWatcher interface immediately displays the prediction result (the sigmoid output) of a pretrained deep neural network via a tooltip view. This additional information facilitates readers in the selection of which news articles to read in detail. Once a user decides to read a news article, the full news story will be made available to them as usual, along with a user feedback section that appears at the bottom of the page. This feedback section was implemented in the form of a button that signals whether the news story was consistent with the context that was provided by the news headline. This process enables the system to gather manual labels on incongruent news articles in the wild, which will be used to train the deep learning model periodically to increase the accuracy and robustness of detection.

To reduce the computational burden of running a deep model, BaitWatcher was implemented as a browser extension that is based on a client-server architecture. After installing the extension on a browser (e.g., Chrome), online users can choose to read news articles on any news platforms and obtain information about the incongruence scores of news headlines before reading the full corresponding body text. As shown in the left side of Figure~\ref{bait_main}, if a reader hovers the mouse cursor over a news headline, BaitWatcher sends an HTTP request from the client to the API server with the hyperlink on the article. The server parses the news content via the Python Newspaper3k library\footnote{https://newspaper.readthedocs.io/}, which uses advanced algorithms with web scrapping to extract and parse online newspaper articles. The parsed content is fed into the pretrained deep learning model to return the incongruence score. The AHDE model with IP was selected as the model since this algorithm realized the best performance in the evaluation experiment. Because the Python Newspaper3k library automatically detects headline and body text, BaitWatcher can be run on any news website. The code and implementation details are publicly available on the GitHub page.\footnote{https://github.com/bywords/BaitWatcher}

\subsection{Focus group interview}

After implementing BaitWatcher, we evaluated its performance in a realistic setting. We conducted a small-scale focus group interview to gain insight into how the provision of additional information about headline incongruence can improve the news reading experience. A total of fourteen participants of ages 20 to 29 were recruited from the second author's institute, all of whom identified themselves as moderate to avid news consumers. All participants said they actively read news articles at least once a week. After hearing a brief introduction to BaitWatcher's functionality, each focus group participant was given 30 minutes to read news articles through the BaitWatcher interface. While BaitWatcher can be deployed on any news website as discussed earlier, we asked the focus group participants to visit a common news portal for finding news~\cite{naver_news} to minimize the effects of distinct media outlets on the perception of headline incongruence. After the 30-minute news reading experience in the lab, we conducted an open interview with each participant. Institutional Review Board had approved this focus group survey and the news assistant experimental design at the second author’s institute (Approval code: \#KH2018-62). 

The open interview included the following questions, which capture the news reading habits of users and quantify the effectiveness of BaitWatcher:

\begin{itemize}[itemindent=2em]
  \item[Q1.] How often do you read news online in a week?
  \item[Q2.] Which category of news are you mostly interested in?
  \item[Q3.] When you are reading news online, how likely are you to read the full story?
  \item[Q4.] Does showing incongruence scores affect the choice of news articles to read?
  
\end{itemize}

\begin{table}[t]
\centering
\begin{tabular}{cccclcc}
\hline
Participant & Gender & Age & Q1 (Freq) & Q2 (Interests) & Q3 (Full story) & Q4 (BaitWatcher) \\\hline
P1 & Male & 24 & 7 days & Politics & Less likely & No\\
P2 & Male & 23 & 7 days & Politics & Less likely & No \\
P3 & Male & 21 & 3-4 days & Politics & Less likely & Yes\\
P4 & Female & 22 & 2 days & Entertainment & Less likely & Yes\\
P5 & Female & 20 & 7 days & Politics & More likely & Yes\\
P6 & Female & 22 & 7 days & Social issues & More likely & Yes\\
P7 & Male & 22 & 5 days & Social issues & Less likely & Yes\\
P8 & Male & 20 & 4 days & Sports & Less likely & Yes\\
P9 & Male & 26 & 3-4 days & Life \& Culture & Less likely & Yes\\
P10 & Male & 26 & 7 days& Economics & More likely & Yes\\
P11 & Female & 24 & 7 days & Entertainment & Less likely & No\\
P12 & Female & 24 & 2 days & IT \& Science& More likely & No\\
P13 & Female & 24 & 7 days  & Sports& Less likely & Yes\\
P14 & Male & 28 & 7 days & Politics & Less likely & Yes\\\hline
\end{tabular}
\vspace{1mm}
\caption{
Participants' information and questionnaire results}
\label{table:questionnaire}
\end{table}

Table~\ref{table:questionnaire} displays the necessary information about the participants and the questionnaire results. Here, we make observations on their reading behaviors and the effects of BaitWatcher in preventing readers from clicking on incongruent headlines. First, as previous studies noted~\cite{english1944study,gabielkov2016social}, a significant degree of participants (78.5\%) reported that they are more likely to consume headlines without reading the full news stories. This skimming behavior may enable them to browse a more extensive set of news stories every day; however, it makes them vulnerable to misleading headlines such as clickbait and incongruent headlines. This result supports the necessity of showing the incongruence score before the user clicks on a headline. 

In response to the question on the effects of BaitWatcher (Q4), ten out of fourteen participants (71.4\%) reported that the use of this interface affected their choices of news articles to read, whereas four participants (28.6\%) responded that they were not influenced by or did not benefit from this web interface. Particularly, \textit{P12}, whose interest is in reading `IT \& Science' news, responded `No' to this question because the participant did not encounter any news stories for which the incongruence score was alarmingly high within the set of news stories that were browsed. Therefore, the user could not experience the benefits of BaitWatcher. The frequency of incongruent headlines is typically low in practice and can vary across topics. Nonetheless, according to a more significant proportion of the participants, having this additional information seems useful and empowering.

Those who answered `Yes' to Q4 reported that BaitWatcher was ``interesting'' and ``effective'' in that they avoided clicking on news headlines with high incongruence scores, as we had hypothesized.  Three participants (P4, P8, and P14) mentioned that they were attracted to such incongruent headlines because they wanted to inspect the articles that BaitWatcher reported to be incongruent to their headlines.

\begin{quote}
    (P4) ``\textit{... At first, I became curious about why certain headlines were labeled as incongruent by BaitWatcher, so I clicked on them and checked how the articles looked ...}''
\end{quote}

The unexpected browsing behaviors support the findings of previous studies on the adverse effects of labeling on the prevention of fake news~\cite{gao2018label,clayton2019real}. From the opinions of two participants (P8 and P9), we identify new potential to mitigate the unnecessary attention that high incongruence scores receive. A possible strategy is to pursue the interpretability of the deep learning models and to present the results as grounds for the high score and how the algorithm works. When an algorithm looks like a black box, users will naturally question its prediction results. Another strategy is to present ample examples of news articles with high incongruence scores in advance of the experiment to facilitate understanding of the participants regarding the general performance of BaitWatcher. 

\begin{quote}
    (P8) ``\textit{... When  BaitWatcher  displayed a high score, it made me wonder, ``why does this headline have such a high score?'' This led me to click the headline and guess the reasoning that the AI used in making the decision...}''
\end{quote}

\begin{quote}
    (P9) ``\textit{... I did not click incongruent headlines because BaitWatcher warned me not to do so. Nonetheless, whenever it (BaitWatcher) showed high scores, I was curious why the AI made such a decision. It may be effective for people to see the internal reasoning process of this AI model ...}''
\end{quote}

Overall, our focus group study demonstrated that the provision of the incongruence score in today's news reading is empowering to users. Web interfaces such as BaitWatcher will not only prevent newsreaders from clicking on news headlines that are likely incongruent to their full linked stories but also gradually build people's trust over time in the model's predictions. Whether one is an avid news reader or not, spending time on incongruent stories is an unpleasant experience in most cases. A headline might be deliberately misleading due to sarcasm, in which case readers could still click on the news article and enjoy reading it even if BaitWatcher's reported score is high. The deep models that are proposed in this work do not yet provide high interpretability, and detection models that are also interpretable could be developed in future studies.

\section{Discussion}

The role of headlines in the news reading experience has been studied in journalism and communication research. News headlines should provide a concise and accurate summary of the main story, thereby enabling readers to decide whether to read the news story~\cite{smith1982comprehensible}. Online social media and the web have become convenient platforms for news consumption. According to Digital News Reports by Reuters Institute~\cite{reuter}, a third fourth of the survey participants replied that they consume news through online media. In contrast to news consumption via traditional outlets such as newspapers, the main content is not shown to readers in online media; only headlines and visual snippets are exposed. Hence, newsreaders are more likely to consume only the news headlines and not the full news stories---a behavior that some refer to as \textit{a shopper of headlines}~\cite{english1944study}. In such environments, if a news headline does not accurately represent the story, it could mislead readers into disseminating false information~\cite{allport1943building,tannenbaum1953effect}, which could lead to pressing social problems. Even though the proportion of incongruent headlines is not large against the numerous news articles that are published each day, an inaccurate impression can percolate through a user's online networks and eventually lead to severe social problems such as polarization, as a previous study similarly discovered in the context of fake news consumption on Twitter~\cite{grinberg2019fake}. This study identified the dangers and problems that are associated with headline-led news reading, and its contributions are three-fold.

First, we release a dataset of 1.7 million news articles that are constructed on the entire articles published in a nation over two years. Due to the sparsity of incongruent headlines in the wild, it requires a considerable amount of time and effort to develop a sizable and balanced dataset via manual annotation. Therefore, previous studies introduced small datasets that are not suitable for training sophisticated models. To address the issue of scalability in the construction of a dataset, we automatically generate incongruent headlines by implanting paragraphs of other articles into the body text. This generation process can be applied to any set of news articles in any language, which will facilitate future studies on the application of data-driven approaches to incongruent headlines.

Second, this study proposes an attention-based hierarchical neural network for the headline incongruence problem. While recurrent neural networks are efficient in modeling sequential information such as text, a body text hinders the propagation of error signals via backpropagation. Thus, inspired by the hierarchical structure of a news article that is composed of paragraphs, we design a hierarchical recurrent network that models word sequences of each paragraph into a hidden state and combines the sequence of the paragraphs through another level of the recurrent neural network. This newly proposed model outperformed baseline approaches with an AUROC of 0.977 on the detection task. 

Third, we implement a lightweight web interface that facilitates the selection by readers of relevant articles to read in a typical scenario of online news consumption in which only headlines are shown. The results of a focus group interview demonstrate the effectiveness of the interface in preventing users from selecting those articles and suggest a future direction for the improvement of deep learning models. Similar to the findings of a recent work~\cite{horne2019rating}, the participants require a high level of interpretability on model predictions, which is not embedded in the proposed models. Following the recent efforts on deep learning~\cite{samek2017explainable}, the development of an interpretable model will help build a high level of trust in machine-based decisions on incongruent headlines, which will be crucial for the utilization of such interfaces in practice.

\subsection{Hierarchical encoders for stance detection}

To further evaluate the generalizability of the deep approaches that are proposed in this paper, we conducted an additional experiment on the FNC-1 dataset~\cite{fnc1}, with the objective of \textit{stance detection}. This problem is similar to the headline incongruence problem in that one must compare the textual relationships between news headlines and the corresponding full content but different in that its target label consists of four separate cases (unrelated, agree, disagree, and discuss). To obtain a similar setting to that of our task, we transformed these four labels into binary labels: ``unrelated'' and ``others''. 

We compared our hierarchical deep learning approaches (AHDE, HRDE, and HRE) with feature-based methods and standard deep learning models. We also considered ensemble models that combine the predictions of XGB and each deep learning model, because an ensemble of XGB and CDE was the winning model of the FNC-1 challenge~\cite{fnc_winner}. XGB outperformed the other single models with an accuracy of 0.9279. Among deep learning models, the AHDE model realized the highest accuracy of 0.8444. The superior performance of XGB over deep approaches might result from insufficient variations among the training instances in the FNC-1 dataset. Even though the training set contains approximately 50 K examples, many news articles that correspond to the independent label were generated from 1,683 original news articles by swapping headlines with one another; thus, 29.7 cases had identical body text.

These reasons might have led the challenge winners to use ensemble models that combine the predictions of feature-based approaches and deep neural networks. The XGB+CDE ensemble model realized the accuracy of 0.9304 and outperformed all the single models. When we combined the predictions of AHDE with XGB, the ensemble model produced the best accuracy of 0.9433. Incorporating the results in Table~\ref{table_accuracy_auroc}, this finding suggests that the proposed hierarchical neural networks effectively learn textual relationships between two texts in contrast to standard approaches. We firmly believe that the highest accuracy of XGB among the single models is due to the limitation of the FNC-1 dataset, as discussed earlier; hence, the ensemble approach may not be necessary if the dataset is sufficiently large for neural network training. According to additional experiments on the dataset that was proposed in this paper, the AHDE model outperformed all combinations of other approaches for the ensemble.

\subsection{Varying perceptions on headline incongruence}

So far, we have treated the incongruent score as an inherent value that is fixed for each news article. We conducted additional surveys using the Amazon Mechanical Turk (MTurk) service to determine whether the general public would agree with the predictions by our models regarding which news articles contain incongruent headlines. We also evaluated whether people's perceptions of incongruence scores vary according to their partisanship, and we hypothesize based on a previous finding that people's knowledge of the veracity of news varies by political stance~\cite{allcott2017social}.

First, we manually gathered news articles from two media outlets. To retrieve as many incongruent news headlines as possible, we selected two media outlets that are considered not trustworthy by common journalistic standards (referring to {mediabiasfactcheck.com}): one was chosen from the conservative media (\textsf{Media A}) and another from the liberal media (we call \textsf{Media B}). We do not reveal these media names, as the choices of media outlets are less of a concern in our study.
Given the definition of an incongruent headline and the incongruent articles that are selected by the model from each media outlet, we asked 100 Amazon Mechanical Turk workers to answer the following question \textit{``Do you think the headline of the above article is incongruent with its body text?''} 

\begin{figure}
  \centering
  \hspace*{-5mm}
   \includegraphics[width=0.8\linewidth]{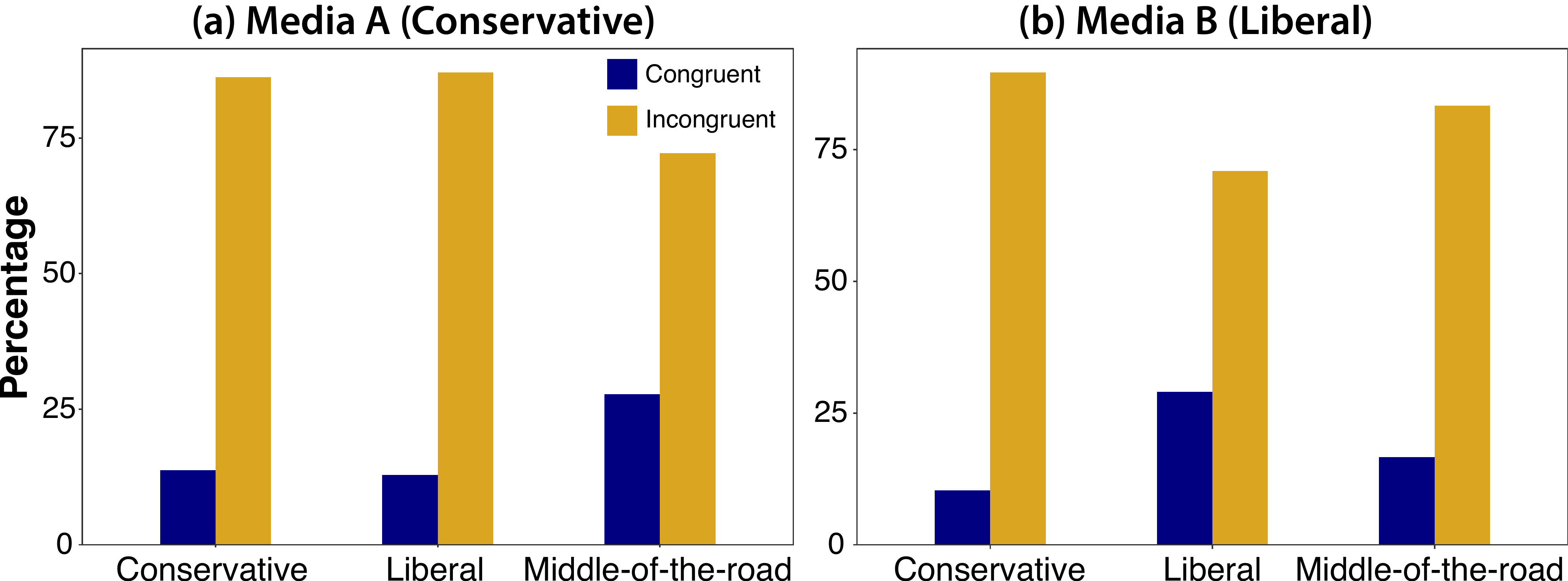}
   \caption{MTurk results indicating political stances of survey participants (the x-axis) and their responses to articles of high incongruence score (the y-axis).}
   \label{fig:result_amt}
\end{figure}

According to Figure~\ref{fig:result_amt}, MTurk workers tend to find that articles with high incongruence scores contain misleading headlines. One exciting trend is the dependence of the perceived incongruence score on individual belief. While nonliberal participants considered news samples from \textsf{Media B} to have a similar level of incongruence to samples from \textsf{Media A}, liberal participants found \textsf{Media B} to be less incongruent. This finding suggests that while our approach is applicable in general scenarios, the perceived incongruence level may be judged differently among news topics (such as politics). News service providers and researchers should be cautious when employing human coders and crowdsourcing workforces to obtain fair labels on misinformation and fake news.

\subsection{Future directions}

A natural extension of this study is the development and improvement of prediction models for detecting news articles with incongruent headlines by incorporating NLP techniques with deep learning approaches. For example, one could apply named entity recognition as a preprocessing step to represent word tokens in an embedding space better. It would also be possible to consider syntactic features in modeling text by developing tree-shaped deep neural architectures that are similar to LSTM-tree~\cite{tai2015improved}.

Another future direction is to devise a learning-based approach for generating headlines. To construct a million-scale dataset for training incongruity detection models, we modified the body text while keeping the original news headline unchanged. While the process has shown to generate a training corpus effectively, researchers could develop an AI agent that rewrites a headline that is incongruent with its body text. While the research on text generation has lagged behind the remarkable achievements in image domains due to the difficulty of handling discrete outputs, future research could be extended from recent studies on controlled generation~\cite{hu2017toward} or cross-alignment style transfer~\cite{shen2017style}.

Beyond the online news domain, this work could lead to new measurements of the incongruence of title and content across other types of online content. The title plays a crucial role in enticing users to click and consume digital content such as blog articles, online videos, and even scientific papers. Similar to the incongruent headline problem, the automatic identification of such incongruent titles of various content will improve people's online experiences. Future researchers could share multiple types of datasets and could develop AI approaches that measure the inconsistency between title and content.

\section{Conclusions}
Here, we study the detection of incongruent headlines that make claims that are unsupported by the corresponding body texts. We release a million-scale data corpus that is suitable for the detection of the misleading headline. We also propose deep neural networks that efficiently learn the textual relationship between headline and body text via a recurrent hierarchical architecture. To further facilitate news reading in practice, we present BaitWatcher, which is a lightweight web interface that presents to readers the prediction results that are based on deep learning models before the readers click on news headlines. The code and implementation details are released for broader use, and we hope this study contributes to the construction of a more trustworthy online environment for reading news.

\section*{Acknowledgements}

This research was supported by Basic Science Research Program through the National Research Foundation of Korea (NRF) funded by the Ministry of Science and ICT (No. NRF-2017R1E1A1A01076400).

\bibliographystyle{splncs04}
\bibliography{reference}

\end{document}